\def\cvprPaperID{12320}
\def\confName{ICCV}
\def\confYear{2023}

\def\paperTitle{CLIP-GCD: Simple Language Guided Generalized Category Discovery}

\def\authorBlock{
    Rabah Ouldnoughi\thanks{Equal contribution} \qquad
    Chia-Wen Kuo\footnotemark[1] \qquad \\
    Georgia Tech \\
    {\tt\small \{rabah.ouldnoughi, albert.cwkuo\}@gatech.edu}
    \and Zsolt Kira \\Georgia Tech\\{\tt\small zkira@gatech.edu}
}

\newif\ifreview 
\newif\ifarxiv \newcommand{\arxiv}{\arxivtrue}
\newif\ifcamera 
\newif\ifrebuttal 

\arxiv %

\pdfoutput=1
\documentclass[10pt,twocolumn,letterpaper]{article}
\ifreview \usepackage[review]{cvpr} \fi
\ifarxiv \usepackage[pagenumbers]{cvpr} \fi
\ifrebuttal \usepackage[rebuttal]{cvpr} \fi
\ifcamera \usepackage{cvpr} \fi

\usepackage{graphicx}
\usepackage{amsmath}
\usepackage{amssymb}
\usepackage{booktabs}

\usepackage{times}
\usepackage{microtype}
\usepackage{epsfig}
\usepackage[table,xcdraw,dvipsnames]{xcolor}
\usepackage{caption}
\usepackage{float}
\usepackage{placeins}
\usepackage{color, colortbl}
\usepackage{stfloats}
\usepackage{enumitem}
\usepackage{tabularx}
\usepackage{xstring}
\usepackage{multirow}
\usepackage{xspace}
\usepackage{url}
\usepackage{subcaption}
\usepackage{xcolor}
\usepackage[hang,flushmargin]{footmisc}

\usepackage[textwidth=1.8cm]{todonotes}
\ifcamera \usepackage[accsupp]{axessibility} \fi

\ifarxiv  \fi

\newcommand{\R}[1]{{%
    \textbf{%
        \ifstrequal{#1}{1}{\textcolor{red}{R#1}}{%
        \ifstrequal{#1}{2}{\textcolor{blue}{R#1}}{%
        \ifstrequal{#1}{3}{\textcolor{magenta}{R#1}}{%
        \ifstrequal{#1}{4}{\textcolor{teal}{R#1}}{%
                           \textcolor{cyan}{R#1}%
        }}}}%
    }%
}}

\usepackage{xr-hyper}

\makeatletter
\newcommand*{\addFileDependency}[1]{
  \typeout{(#1)}
  \@addtofilelist{#1}
  \IfFileExists{#1}{}{\typeout{No file #1.}}
}

\makeatother

\usepackage[pagebackref,breaklinks,colorlinks]{hyperref}
\usepackage[capitalize]{cleveref}
\crefname{section}{Sec.}{Secs.}
\crefname{table}{Table}{Tables}
\crefname{figure}{Fig.}{Figs.}

\begin{document}
\title{\paperTitle}
\author{\authorBlock}
\maketitle

\begin{abstract}
Generalized Category Discovery (GCD) requires a model to both classify known categories and cluster unknown categories in unlabeled data. Prior methods leveraged self-supervised pre-training combined with supervised fine-tuning on the labeled data, following by simple clustering methods. In this paper, we posit that such methods are still prone to poor performance on out-of-distribution categories, and do not leverage a key ingredient: Semantic relationships between object categories. We therefore propose to leverage multi-modal (vision and language) models, in two complementary ways. First, we establish a strong baseline by replacing uni-modal features with CLIP, inspired by its zero-shot performance. 
Second, we propose a novel retrieval-based mechanism that leverages CLIP's aligned vision-language representations by mining text descriptions from a text corpus for the labeled and unlabeled set.
We specifically use the alignment between CLIP's visual encoding of the image and textual encoding of the corpus to retrieve top-k relevant pieces of text, and incorporate their embeddings to perform joint image+text semi-supervised clustering. 
We perform rigorous experimentation and ablations (including on where to retrieve from, how much to retrieve, and how to combine information), and validate our results on several datasets including out-of-distribution domains, demonstrating state-of-art results. 
On the generic image recognition datasets, we beat the current state of the art (XCon \cite{Fei2022XConLW}) by up to 6.7\% on all classes, up to 2.0\% on known classes, and 11.6\% on average over unknown classes,  and on fine-grained datasets up to 14.3\% on average over all classes, and up to 10.7\% on average over unknown classes.
\end{abstract}
\section{Introduction}
\begin{figure}[t]
\captionsetup{singlelinecheck = false, justification=justified}
\begin{center}
    \includegraphics[width=0.49\textwidth]{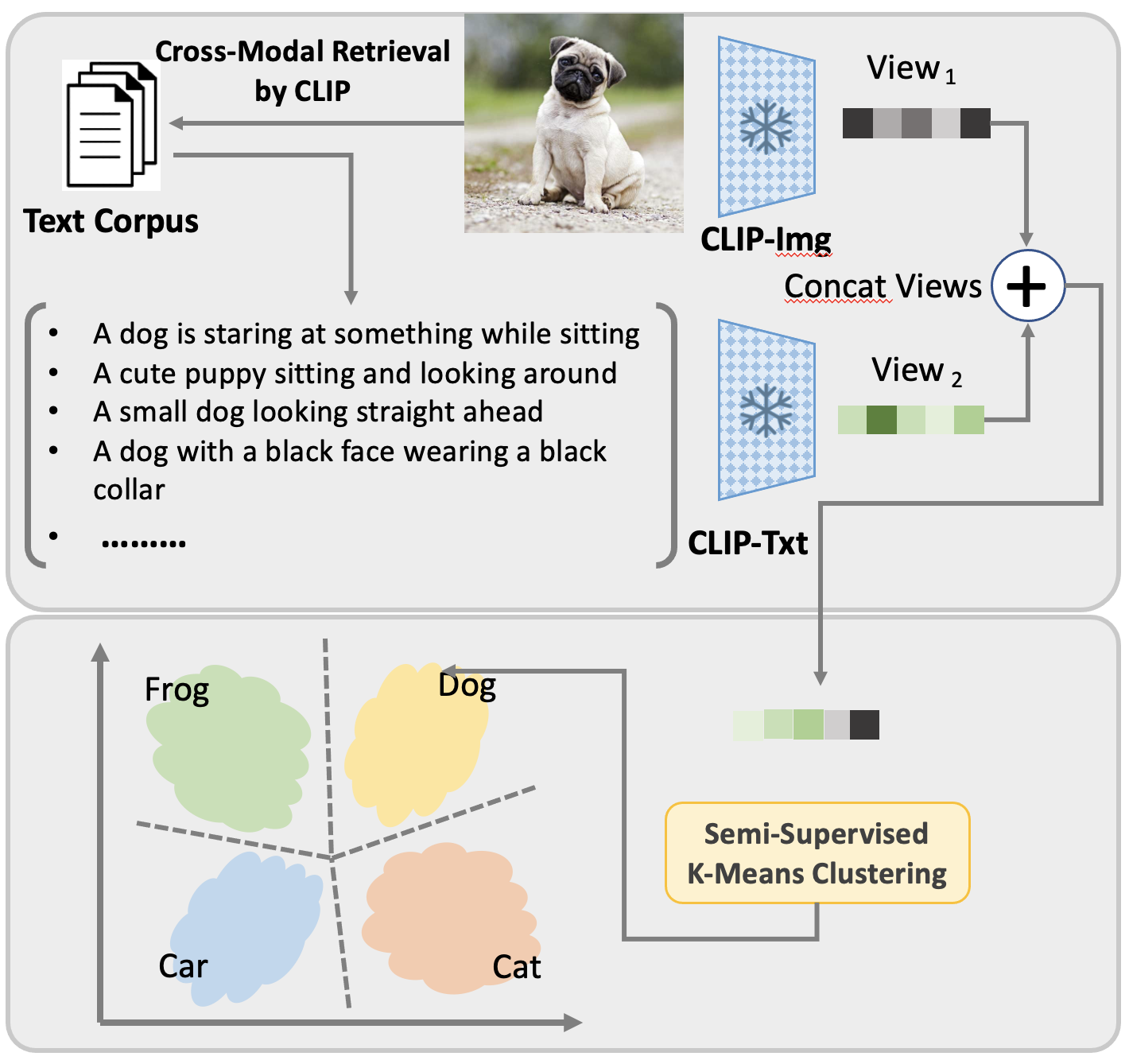}
\end{center}
\caption{
We propose a novel retrieval-based clustering mechanism to improve the representation of the input image for generalized category discovery (GCD).
\textbf{\textit{(Top)}} We first leverage CLIP’s aligned vision-language representations to retrieve a set of highly relevant text descriptions from a large text corpus using the input image as a query.
To further leverage CLIP's large-scaled pre-trained representation, the input image, and its retrieved texts are encoded by a \textbf{frozen} CLIP image and text encoder into a set of feature encodings.
\textbf{\textit{(Bottom)}} Given the concatenated text and image views, we adopt the semi-supervised \textit{k}-means clustering to cluster features into seen and unseen classes.
}
\label{fig:figure1}
\end{figure}

\looseness=-1 Despite tremendous progress in computer vision, a number of limitations remain. One important limitation is that all categories must be known or annotated \textit{a-priori}. In other words, deep learning cannot \textit{discover} new categories not reflected in the original training set. This limits applicability to a range of problem domains, including self-driving cars or personal devices, where new categories will inevitably appear often without annotation or even knowledge of which categories are known and which are not. The setting of Novel Category Discovery (NCD)~\cite{Hsu2018LearningTC} tackles this problem of discovering new categories in unlabeled data, by leveraging pre-training or auxiliary training. Recently, Generalized Category Discovery (GCD) \cite{Vaze2022GeneralizedCD} was formalized to make the setting more realistic, where the goal is to jointly discover unknown categories and classify known categories within the unlabeled data.
This setting is related to semi-supervised learning but does not assume we know all of the classes in the data \cite{Chapelle2006SemiSupervisedL, Sohn2020FixMatchSS}. 

The state-of-the-art methods for this setting utilize self-supervised image pre-training (e.g. DINO~\cite{Caron2021EmergingPI}) as auxiliary information used to encode the images, after which simple clustering is performed~\cite{Vaze2022GeneralizedCD}. 
However, even though self-supervised feature learning can show some out-of-domain generalization \cite{Khalid2022RODDAS, Hendrycks2019UsingSL}, it is still a difficult challenge as the features may not be relevant to entirely new categories. 

In this paper, we posit that a key missing element to improve such generalization is a more effective encoding of the semantic relationships between object categories. Recently, aligned multi-modal (vision and language) models have been shown to give a remarkable boost in the generalization of visual learning, especially when scaled up \cite{Furst2021CLOOBMH, Jia2021ScalingUV, Li2022SupervisionEE, Radford2021LearningTV}. 
These models are learned via alignment of visual and language embeddings through large-scale constrastive training of paired image-text data \cite{Radford2021LearningTV}. Such methods have demonstrated a potential for learning open-world visual concepts, since the textual alignment forces visual features to be nearby similar concepts, and hence new categories can be well-placed in the feature space by the visual encoder.

\looseness=-1 Given the strong zero-shot results of such models, we, therefore, propose to first \textit{replace} the uni-modal image encoder with one trained in a multi-modal fashion (CLIP~\cite{Radford2021LearningTV}). By itself, this simple modification yields significant performance gains, beating all of the current state of the art. Hence, this setting can serve as a simple, but extremely strong, baseline. 

However, in just replacing the visual encoder, we discard the text branch of the multi-modal model and thus fail to fully leverage the joint vision-and-language (VL) embedding and its zero-shot generalizability.
Furthermore, despite significant gains, the visual encoder from a multi-modal model can still perform poorly when the visual concepts are not well-represented in their training data and are somewhat out-of-distribution.

In this paper, we propose to \textit{augment} the visual embeddings with \textit{retrieved} textual information. This allows us to better leverage the joint VL embedding and the text encoder as well as provide the ability to extend the contextual knowledge available for clustering unknown and potentially out-of-domain categories and images. Specifically, inspired by prior image captioning works~\cite{kuo2022beyond}, given an image, we retrieve the top-\textit{k} most relevant text from a large text corpus~\cite{Gong2014ImprovingIE, HodoshMicah2013FramingID} (which could be from the multi-modal training set itself). We specifically use the alignment between CLIP's visual encoding (of the image) and textual encoding (pre-indexed for the text corpus). Our key hypothesis is that such pieces of text, and their encodings, can provide valuable contextual clues for clustering unseen categories.
The retrieved top-\textit{k} text are encoded by CLIP's text encoder, are mean-pooled, and then concatenated with the CLIP's visual encoding as the final multi-modal representation for clustering.

We show that our proposed method substantially outperforms the established state of the art across a number of datasets. We specifically expand the set of datasets to include out-of-domain data, DomainNet (a domain adaptation dataset), and Flowers102, a generic image recognition dataset. We perform extensive analysis of what corpus to retrieve from, how much to retrieve, and how to combine (or pool) the resulting embeddings. Crucially, we demonstrate in our ablation studies that the \textbf{combination of our two ideas} (using CLIP and retrieving contextual information) is needed to yield strong state of art results. This is because combined clustering of \textit{aligned} embeddings is significantly more effective than clustering individual image and textual embeddings that are not aligned. 

In summary, we make the following contributions:
\begin{itemize}
  \item We propose a simple but extremely effective baseline for GCD, utilizing CLIP image encodings rather than uni-modal pre-trained ones. 
  \item We further propose a cross-modal retrieval module by leveraging the cross-modal joint embedding space of CLIP to retrieve a set of contextual text descriptions for unlabeled data containing seen and unseen categories. 
  \item We perform extensive experimentation, including on more challenging out-of-distribution datasets, demonstrating Significant improvements over the state-of-art (and even our strong baseline) alongside rigorous quantitative and qualitative analysis of our approach.
\end{itemize}
\label{sec:intro}

\section{Related Work}
\subsection{Novel Category Discovery (NCD)}
NCD is a relatively nascent field, first proposed as ``cross-task transfer'' where learning on labeled data can be transferred to clustering of unseen categories (disjoint from the labeled set) in unlabeled data~\cite{Hsu2018LearningTC, Hsu2019MulticlassCW}. 
Several methods have been developed to tackle this task. \cite{Hsu2018LearningTC, Hsu2019MulticlassCW} use a pair-wise siamese network trained on labeled data and apply it to train a clustering network on unlabeled data. Subsequent works improved upon this via a specialized deep clustering approach~\cite{Han2019LearningTD}. In RankStat \cite{Han2020AutomaticallyDA,Han2022AutoNovelAD}, a three-stage pipeline is deployed: The model is trained with self-supervision initially on all data for representation learning, then fine-tuned on labeled data to capture higher-level semantic knowledge, and finally ranking statistics are used to transfer knowledge from the labeled to unlabeled data. \cite{Zhong2021NeighborhoodCL} presents a contrastive learning approach, generating hard negatives by mixing labeled and unlabeled data in the latent space. UNO \cite{Fini2021AUO} introduces a unified cross-entropy loss, jointly training a model on labeled and unlabeled data by trading pseudo-labels from classification heads.
Our work builds on top of a new and more realistic setting named Generalized Category Discovery (GCD) \cite{Vaze2022GeneralizedCD} where the unlabeled samples can come both from seen and unseen classes. The original GCD method performed \textit{k}-means based clustering of DINO embeddings, while recent developments such as XCon~\cite{Fei2022XConLW} have improved those results through additional contrastive training. In our paper, we focus on leveraging multi-modal models in several ways, which is orthogonal to such improvements. We also demonstrate superior results compared to all of the current published state of the art.   
\subsection{Unsupervised Clustering}
Clustering has a long history and has long been studied by the machine-learning community. The task is to automatically partition an unlabeled dataset into different semantic groups without access to information from a labeled set. To tackle this task, several shallow \cite{Arthur2007kmeansTA, MacQueen1967SomeMF,ZelnikManor2004SelfTuningSC} and deep learning \cite{hsu2015neural, Chang2017DeepAI,Rebuffi2021LSDCLS,VanGansbeke2020LearningTC,Yang2017TowardsKS} approaches have been proposed. The deep learning-based methods can be roughly divided into two types, the first of which uses the pairwise similarity of samples to generate pseudo-labels for clustering and the second of which uses neighborhood aggregation to coalesce similar samples while at the same time pushing apart dissimilar samples, achieving a clustering effect. Such advanced clustering methods could be added to our approach, though we focus on improving the underlying feature space such that simple clustering methods can be used. 

\subsection{Self-Supervised and Multi-Modal Pre-Training}
\looseness=-1 Self-supervised learning has advanced rapidly over the years. Some methods leverage contrastive learning, often across augmented copies of the unlabeled image, by breaking symmetry e.g. via projection heads~\cite{chen2020simple} or teacher-student training where the teacher comes from some version of the student (e.g. an exponential moving average of the student over the iterations)~\cite{grill2020bootstrap}. 
Recently, the advent of Vision Transformers (e.g. ViT)~\cite{dosovitskiy2020image}, which have significantly more flexibility and capacity, has enabled these methods both to scale (i.e. further improve) with larger unlabeled datasets~\cite{Caron2021EmergingPI} as well as provide unique opportunities for new mechanisms such as masking~\cite{he2022masked}. Besides unlabeled data, multi-modal methods leverage image-text pairs mined from the web. Again, methods such as contrastive learning can be used to push image and text embeddings together (when paired) or apart (when not). Methods such as CLIP~\cite{Radford2021LearningTV}, which do this across very large datasets, have shown impressive zero-shot performance. All of these methods are relevant to the GCD problem, as category discovery benefits from better representations (with self-supervised learning having nice properties out-of-distribution) and zero-shot classification is a similar problem except that in GCD the collection of unlabeled data is available. Further, our method explicitly leverages the alignment between image and text encoders in multi-modal models to better cluster unlabeled data.
\label{sec:related}

\section{Method}
\begin{figure*}[t]
\begin{center}
\captionsetup{singlelinecheck = false, justification=justified}
\includegraphics[width=0.73\textwidth]{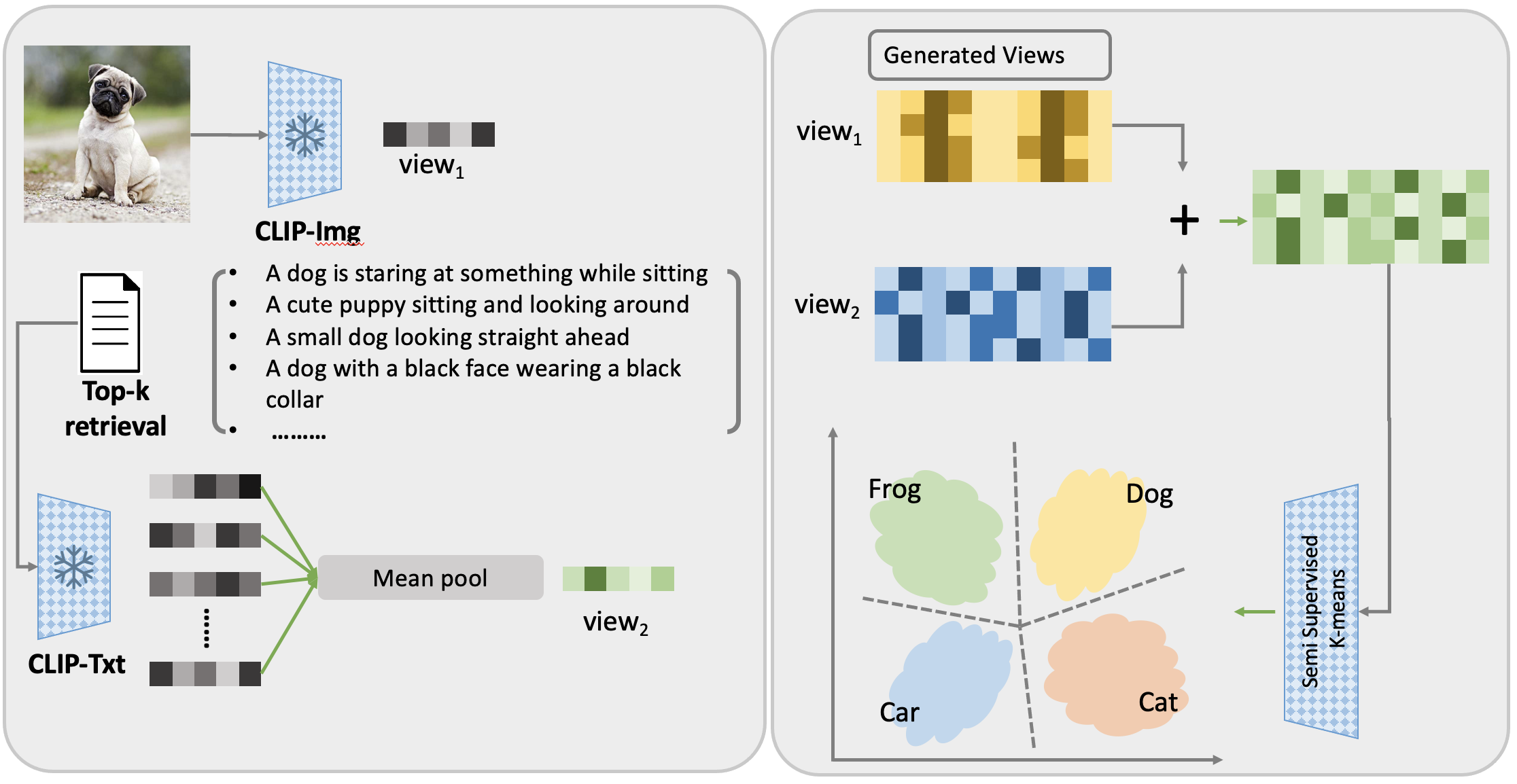}
\end{center}
\caption{Model Architecture. \textit{In stage I (left), we propose a cross-modal retrieval module to retrieve a set of contextual text descriptions for the labeled and unlabeled data, generate a view from pooled sentence embedding as complementary information for clustering. In Stage II (right), we concatenate the image view and the text view and use semi-supervised \textit{k}-means clustering to group seen and unseen classes.}}
\label{fig:architecture}
\end{figure*}
In this section, we first introduce the notations and definitions of GCD \cite{Vaze2022GeneralizedCD}. Then, we explain how to use CLIP in GCD and introduce our method to tackle this task.
 \subsection{Problem Setup of GCD}
As formalized in \cite{Vaze2022GeneralizedCD}, dataset $\mathcal{D}$ consists of two parts, labeled dataset  $\mathcal{D_{L}}=\{(\mathbf{x}_{i},y_{i})\}^N_{i=1} \in \mathcal{X} \times \mathcal{Y_{L}} $ and unlabeled dataset $\mathcal{D_{U}}=\{(\mathbf{x_{i},y_{i}})\}^M_{i=1} \in \mathcal{X} \times \mathcal{Y_{U}}$, where $\mathcal{Y_{L}} \subset \mathcal{Y_{U}}$ which is distinct from NCD \cite{Han2019LearningTD} that assumes $\mathcal{Y_{L}} \cap \mathcal{Y_{U}}=\emptyset$. The goal is to learn a model to group the instances in $\mathcal{D_{U}}$ based on information from $\mathcal{D_{L}}$. Taking advantage of the recent advances in vision transformers and their remarkable performance in various visual recognition tasks specifically for self-supervised representation learning \cite{Caron2021EmergingPI}, Vaze \textit{et al.} \cite{Vaze2022GeneralizedCD} devise a two-stage training pipeline for the GCD task. First, for representation learning, they jointly fine-tune the representation by performing supervised contrastive learning on the labeled data and unsupervised contrastive learning on all the data.

Let $\mathbf{x}_{i}$ and $\mathbf{x}_{i}'$ be two views with random augmentations of the same image in a mini-batch $\mathit{B}$. The unsupervised contrastive loss is stated as:

\begin{center}
\label{eq:1}
$\mathcal{L}^{u}_{i}=- \text{log}\frac{\text{exp}(\mathbf{z}_{i}\cdot\mathbf{z}_{i}'/\tau)}{\sum{_n} 1_{[n \neq i]}\text{exp}(\mathbf{z}_{i}\cdot\mathbf{z}_{n}'/\tau)'}$
\end{center}
where $\mathbf{z}_{i}=h(f(x_{i}))$ is the feature extracted by a backbone $f(\cdot)$ on the input image $\mathbf{x}_{i}$ and projected to do the embedding space via a projection head $h(\cdot)$, $\mathbf{z}_{i}'$ is the feature from another view of the input image $\mathbf{z}_{i}'$.

The supervised contrastive loss is stated as
\begin{center}
\label{eq:2}
$\mathcal{L}^{s}_{i}=- \frac{1}{|\mathcal{N}(i)|}\sum\limits_{{q\in\mathcal{N}(i)}} \text{log}\frac{\text{exp}(\mathbf{z}_{i}\cdot\mathbf{z}_{q}/\tau)}{\sum{_n} 1_{[n \neq i]}\text{exp}(\mathbf{z}_{i}\cdot\mathbf{z}_{n}/\tau)'}$
\end{center}
where $\mathcal{N}(i)$ denotes the indices of other images having the same label as $\mathbf{x}_{i}$ in the mini-batch 
$\mathit{B}$. Then, the final objective is the combination of the two losses: 
\begin{center}
\label{eq:3}
    $\mathcal{L}^{t}=(1-\lambda)\sum\limits_{i\in\mathcal{B}_{\mathcal{L}}\cup\mathcal{B}_{\mathcal{U}}}\mathcal{L}^{u}_{i}+ \lambda \sum\limits_{i\in B_{\mathcal{L}}} \mathcal{L}^{s}_{i}$
\end{center}
where $\lambda$ is a weight factor and $\mathcal{B_{\mathcal{L}}}$, $\mathcal{B_{\mathcal{U}}}$ are mini-batches for labeled and unlabeled images respectively. For label assignments, a semi-supervised  \textit{k}-means is proposed, where the overall procedure is similar to \textit{k}-means \cite{MacQueen1967SomeMF} However, there is a significant distinction in that semi-supervised \textit{k}-means takes into account the labeled data in $\mathcal{D_{L}}$ during the computation of cluster assignment in each step. This means that the samples with labels will always be assigned to the correct cluster, irrespective of their distance to the nearest cluster centroids.

\subsection{Our Approach}
By combining both textual and visual information, language-image models can achieve improved performance in a wide range of tasks, so we propose to leverage CLIP's zero-shot ability and multi-modal aligned encoders for this setting, and then propose a retrieval-based augmentation. 

\subsubsection{Using CLIP in General Category Discovery}
\begin{figure}[h!]
\captionsetup{singlelinecheck = false, justification=justified}
\begin{center}
    \includegraphics[width=0.35\textwidth]{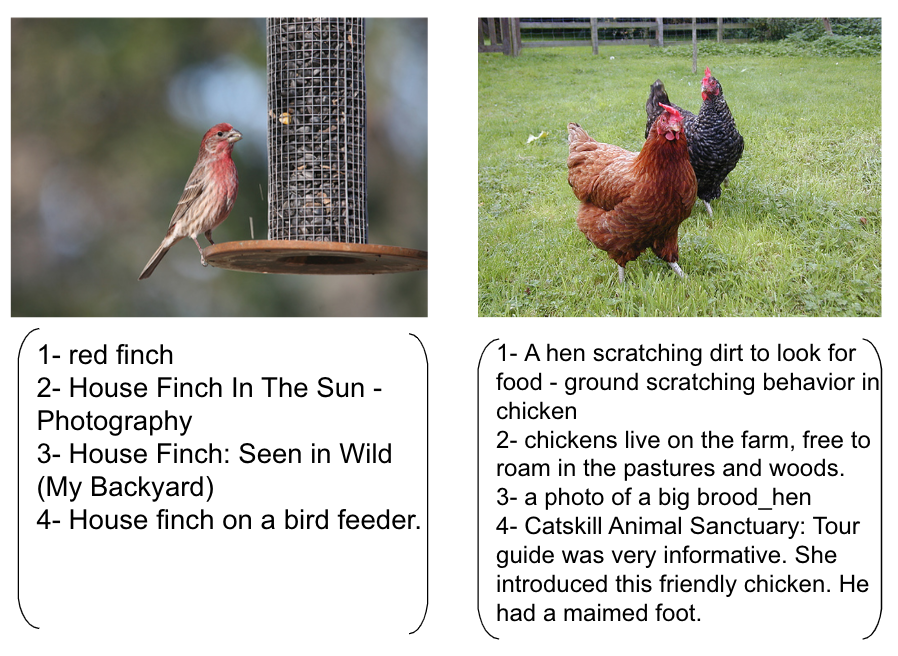}
\end{center}
\caption{\textit{A sample of retrieved top-4 most relevant text descriptions from Conceptual Captions (3M) for an image from ImageNet dataset.}}
\label{fig:figure3}
\end{figure}

We propose to tackle the GCD task by leveraging the cross-modal joint embedding from CLIP \cite{Radford2021LearningTV}. The CLIP model has two branches: the image branch CLIP-Image and the text branch CLIP-Text that encode image and text into a global feature representation, respectively. CLIP is trained on large-scale image and text pairs \textit{s.t.} paired image and text are pushed together in the embedding space while unpaired ones are pulled apart. Please refer to Figure \ref{fig:architecture} for the overall architecture. To improve the representation of our data, specifically for both labeled and unlabelled data, we refine the representation by combining two techniques: supervised contrastive learning on the labeled data and unsupervised contrastive learning on all data. We do this by finetuning the representation on our target data simultaneously. CLIP learns image representation by contrasting them with the representations of text description of the image, such as \textit{``A photo of a \{\text{class name}}\}''. 
The text description is called \textit{prompt}, and its design is vital in enhancing CLIP's performance. However, the unlabeled data contains unseen categories, and we do not have a list of them to use for prompts. 
As a result, inspired by recent works in image captioning~\cite{kuo2022beyond}, for the labeled and unlabeled set, we propose to mine a set of text descriptions providing complementary information to the input image from a text corpus. The  key hypothesis is that such contextual information, provided as additional ``views'' of the image, can significantly aid in clustering.
To that end, we propose to generate text descriptions for an image as shown in Figure \ref{fig:figure3} containing details and information of the input image to be mapped into the feature space. 
Training a separate captioning model to generate text descriptions might be expensive and nontrivial, so for each labeled and unlabeled image, we retrieve the top-\textit{k} most relevant descriptions from a text corpus, turning this problem into a cross-model retrieval one, which we  describe as follows.

\textbf{Description database} The description database is an organized collection of textual descriptions relevant to an image, and we select the top-\textit{k} most pertinent ones. 
Several options can be used, and we show results across annotations from databases such as Conceptual Captions (3M) \cite{Sharma2018ConceptualCA}, Conceptual Captions (12M) \cite{Changpinyo2021Conceptual1P}, MS Coco \cite{Lin2014MicrosoftCC}, and LION \cite{Schuhmann2021LAION400MOD}. 
We don’t perform any rigorous processing and simply collect all the captions. 

\textbf{Text description retrieval} Given a query of an image, the goal is to retrieve the top-\textit{k} most relevant text descriptions from the description database. To this end, we propose to exploit the cross-modal joint embedding from CLIP \cite{Radford2021LearningTV} for this cross-modal retrieval task. Specifically, we use CLIP-Text to encode all the descriptions in the description database as the search key. The image is encoded by CLIP-Image into a query. We then search in the description database for the text descriptions with the top-\textit{k} highest cosine similarity scores. Some examples of the top-4 results are shown in Figure \ref{fig:figure3}.

\textbf{Multi-view generation for clustering} The general approach for our feature vector extraction and view generation framework is illustrated in Figure  \ref{fig:architecture} (Stage I). Given an image and a set of text descriptions, an image view (feature vector) is generated by encoding it using the CLIP image encoder, then using the CLIP text encoder, we encode the set of text descriptions, pool embeddings, and generate a view (sentence embedding) using mean pooling. Finally, the feature vectors of the image and text (views) are concatenated and projected into CLIP latent space, and clustering is performed directly in it. \\

\textbf{Label assignment with semi-supervised k-means clustering}
Given the image view and the text view we concatenate the feature vectors and apply semi-supervised \textit{k}-means clustering following \cite{Vaze2022GeneralizedCD} to group the unlabeled data into seen and
unseen classes. The semi-supervised \textit{k}-means is a transformation of the traditional k-means method into a constraint-based algorithm, where the number of clusters \textit{k} is assumed known. This will involve requiring that the $\mathcal{D}_{\mathcal{L}}$ data instances are assigned to their appropriate clusters based on their ground-truth labels. The first set of centroids \textbar$\mathcal{Y}_{\mathcal{L}}$\textbar for $\mathcal{D}_{\mathcal{L}}$ in semi-supervised k-means are obtained using actual class labels. The second set of centroids for the additional number of new classes \textbar$\mathcal{Y}_{\mathcal{U}}$\textbackslash$\mathcal{Y}_{\mathcal{L}}$\textbar are obtained from $\mathcal{D}_{\mathcal{U}}$ using k-means++\cite{Arthur2007kmeansTA}, but only within the constraint of $\mathcal{D}_{\mathcal{L}}$ centroids. During the process of updating and assigning centroids, instances from the same class in $\mathcal{D}_{\mathcal{L}}$ are always grouped together, whereas instances in $\mathcal{D}_{\mathcal{U}}$ can be assigned to any cluster based on their distance to various centroids. After the algorithm converges, each instance in $\mathcal{D}_{\mathcal{U}}$ can be given a cluster label.

\section{Experiments}
\subsection{Model architecture details} CLIP \cite{Radford2021LearningTV} has two encoders, CLIP-Image and CLIP-Text which are pre-trained transformer models for image and text. CLIP-Text is a base transformer model consisting of 12 layers, a hidden size of 768, and the final linear projection layer produces a representation vector of size 512. 
CLIP-Image is a hybrid ViT-Base model (which is the same as the DINO-trained model used for a fair comparison) consisting of 12 stacked layers, with a convolutional layer in the beginning for feature extraction. For a given image, a total of 49 embedding vectors with a hidden size of 768 are generated, and to match the output of the CLIP-Text encoder; the output hidden state is projected from 768 to 512 dimensions. We fine-tune the last block of the vision transformer starting with a learning rate of 5e-5 decaying it over time using a cosine annealed schedule. We train the model for 100 epochs using batches of size 128 and set the value of $\lambda$ to 0.25 in the loss function (Eq. (\ref{eq:3})). Tuning and testing is done on a separate validation set to select the best hyperparameters. 

\begin{table*}[t]
\caption{Comparative results on generic image recognition datasets}
\label{tab:resultTableCL}
\centering
\begin{tabular}{lcccccccccccccccccc}
\toprule
 & \multicolumn{3}{c}{CIFAR10} &  & \multicolumn{3}{c}{CIFAR100} &  & \multicolumn{3}{c}{ImageNet-100} & & \multicolumn{3}{c}{Flowers-102}\\ \cline{2-4} \cline{6-8} \cline{10-12} \cline{14-16}
Classes & All & Old & New &  & All & Old & New &  & All & Old & New &  & All & Old & New\\ \hline
RankStats+ \cite{Han2022AutoNovelAD} & 46.8 & 19.2 & 60.5 &  & 58.2 & 77.6 & 19.3 &  & 37.1 & 61.6 & 24.8 &  & - & - & -\\
UNO+ \cite{Fini2021AUO} & 68.6 & \textbf{98.3} & 53.8 &  & 69.5 & 80.6 & 47.2 &  & 70.3 & 95.0 & 57.9 &  & - & - & -\\
GCD \cite{Vaze2022GeneralizedCD} & 91.5 & 97.9 & 88.2 &  & 73.0 & 76.2 & 66.5 &  & 74.1 & 89.8 & 66.3 &  & 74.1 & 82.4 & 70.1\\
GCD w/ CLIP & 95.9 & 97.0 & 95.8 &  & 84.2 & 83.1 & 82.3 &  & 79.3 & 94.6 & 71.1 &  & 67.8 & 82.3 & 60.5\\ 
XCon \cite{Fei2022XConLW} & 96.0 & 97.3 & 95.4 &  & 74.2 & 81.2 & 60.3 &  & 77.6 & 93.5 & 69.7 &  & - & - & -\\ \hline
Ours & \textbf{96.6} & 97.2 & \textbf{96.4} &  & \textbf{85.2} & \textbf{85.0} & \textbf{85.6} &  & \textbf{84.0} & \textbf{95.5} & \textbf{78.2}  &  & \textbf{76.3} & \textbf{88.6} & \textbf{70.2} \\ \bottomrule
\end{tabular}
\end{table*}

\subsection{Datasets \& Evaluation}
We evaluate the performance of our method on both generic image classification and fine-grained datasets. Following \cite{Vaze2022GeneralizedCD}, we selected CIFAR-10/100 \cite{Krizhevsky2009LearningML}, ImageNet-100 \cite{Deng2009ImageNetAL}, and Flowers102 \cite{Nilsback08} as the generic image classification datasets. We use CUB-200 \cite{Wah2011TheCB}, Stanford Cars \cite{Krause20133DOR}, and FGVC-Aircraft \cite{Maji2013FineGrainedVC} as fine-grained datasets. We also experiment with a challenging domain adaptation dataset DomainNet (Sketch) \cite{peng2019moment}. We split the training data into two parts, a labeled dataset and an unlabeled dataset by dividing all classes equally into seen classes and unseen ones, then sampling 50\% images from the seen classes as unlabeled data so that the unlabeled set $\mathcal{D}_{\mathcal{U}}$ contains images from both seen classes and unseen classes, while the labeled set only contains seen classes. The splits are summarized in Table \ref{tab:datasetssplits}.

\begin{table}[h!]
\resizebox{\columnwidth}{!}{%
\begin{tabular}{lcccc}
\toprule
 & CIFAR10 & CIFAR100 & CUB-200 & SCARS \\ \hline
\textbar$\mathcal{Y_L}$\textbar & 5 & 80 & 100 & 98 \\
\textbar$\mathcal{Y}_{\mathcal{U}}$\textbar & 10 & 100 & 200 & 196 \\ \hline
\textbar$\mathcal{D_L}$\textbar & 12.5k & 20k & 1.5k & 2.0k \\
\textbar$\mathcal{D}_{\mathcal{U}}$\textbar & 37.5k & 30k & 4.5k & 6.1k \\ \hline
 & ImageNet-100 & DomainNet(Sketch) & Flowers-102 & FGVC-Aircraft \\ \hline
\textbar$\mathcal{Y_L}$\textbar & 50 & 172 & 51 & 50 \\
\textbar$\mathcal{Y}_{\mathcal{U}}$\textbar & 100 & 345 & 102 & 100 \\ \hline
\textbar$\mathcal{D_L}$\textbar  & 31.9K & 10.1k & 255 & 1.7k \\
\textbar$\mathcal{D}_{\mathcal{U}}$\textbar & 95.3k & 38k & 765 & 5k \\ \bottomrule
\end{tabular}%
}
\caption{
Our dataset splits in the experiments. (\textbar$\mathcal{Y_L}$\textbar,\textbar$\mathcal{Y}_{\mathcal{U}}$\textbar) correspond to the number of classes in the labeled and unlabeled sets respectively. (\textbar$\mathcal{D_L}$\textbar,\textbar$\mathcal{D}_{\mathcal{U}}$\textbar) is the number of images for each set.
}
\label{tab:datasetssplits}
\end{table}

\looseness=-1 \textbf{Evaluation Metric} To measure the performance of our model, we use the clustering accuracy (ACC) defined below. 
\begin{center}
$ACC = \underset{p \in P(\mathcal{Y}_{\mathcal{U}})}{\max}\frac{1}{N}\sum\limits_{i=1}^N 1\{y_{i} = p(\hat{y}_{i})\}$
\end{center}
where $\mathcal{P}$ is the set of all permutations that matches the model’s predictions $\hat{y}_{i}$ and the ground truth labels $y_{i}$ using the Hungarian method \cite{Kuhn2010TheHM} and $N$ is the total number of images in the unlabeled set. Following \cite{Vaze2022GeneralizedCD}, we use the metric on three different sets, \textit{'All’} which refers to the entire unlabeled set $\mathcal{D}_{\mathcal{U}}$, \textit{‘Old’} referring to instances in $\mathcal{D}_{\mathcal{U}}$ belonging to classes in $\mathcal{Y}_{\mathcal{L}}$, and \textit{‘New’} referring to instances in $\mathcal{D}_{\mathcal{U}}$ belonging to $\mathcal{Y}_{\mathcal{U}}$ \textbackslash $\mathcal{Y}_{\mathcal{L}}$.

\begin{table*}[t]
\caption{Comparative results on SSB \cite{Oza2019C2AECC} and DomainNet \cite{peng2019moment} }
\label{tab:resultTableFG}
\centering
\begin{tabular}{lccccccccccccccc}
\toprule
 & \multicolumn{3}{c}{Stanford Cars} &  & \multicolumn{3}{c}{FGVC-Aircraft} &  & \multicolumn{3}{c}{DomainNet (Sketch)} &  & \multicolumn{3}{c}{CUB-200} \\ \cline{2-4} \cline{6-8} \cline{10-12} \cline{14-16}
Classes & All & Old & New &  & All & Old & New &  & All & Old & New &  & All & Old & New\\ \hline
RankStats+  \cite{Han2022AutoNovelAD}& 28.3 & 61.8 & 12.1 &  & 26.9 & 36.4 & 22.2 &  & - & - & - &  & 33.3 & 51.6 & 24.2\\
UNO+ \cite{Fini2021AUO}& 35.5 & 70.5 & 18.6 &  & 40.3 & 56.4 & 32.2 &  & - & - & - &  & 35.1 & 49.0 & 28.1\\  
GCD \cite{Vaze2022GeneralizedCD} & 39.0 & 57.6 & 29.9 &  & 45.0 & 41.1 & 46.9 &  & 45.2 & 50.4 & 43.3 &  & 51.3 & 56.6 & 48.7\\ 
GCD w/ CLIP & 62.8 & 85.2 & 52.0 &  & 43.7 & 52.8 & 39.2 &  & 52.7 & 74.2 & 43.7 &  & 59.7 & 76.1 & 51.5\\
XCon \cite{Fei2022XConLW} & 40.5 & 58.8 & 31.7 &  & 47.7 & 44.4 & \textbf{49.4} &  & - & - & - &  & 52.2 & 54.3 & 51.0\\ \hline
Ours & \textbf{70.6} & \textbf{88.2} & \textbf{62.2} &  & \textbf{50.0} & \textbf{56.6} & 46.5&  & \textbf{55.2} & \textbf{75.5} & \textbf{47.4} &  & \textbf{62.8} & \textbf{77.1} & \textbf{55.7}\\ \bottomrule
\end{tabular}
\end{table*}

\subsection{Comparison with the State-of-the-Art}
\looseness=-1 We start by comparing our method with the SOTA methods on both generic image classification, fine-grained image classification, and domain adaptation benchmarks. RankStats+ \cite{Han2022AutoNovelAD} and UNO+ \cite{Fini2021AUO} are two methods modified from two competitive baselines for NCD and adopted to the GCD setting. XCon \cite{Fei2022XConLW} is a method that targets fine-grained datasets in the GCD setting, lastly, GCD w/ CLIP is our proposed use of the GCD method with CLIP image encoder in lieu of DINO. The results on generic image recognition benchmarks are shown in Table \ref{tab:resultTableCL}. On all the datasets we experimented with, our method shows the best performance across most of the categories, often improving upon previous works with large margins. On ImagetNet-100, CIFAR100, and Flowers102, our method outperforms the other methods on all subsets '\textit{All}', '\textit{Old}', and '\textit{New}', reinforcing the idea that our dual usage of multi-modal models boosts performance compared to vision only models. On the fine-grained image classification benchmarks, our results are presented in Table \ref{tab:resultTableFG}. We show the best performance of our method on all categories '\textit{All}', '\textit{Old}', and '\textit{New}' for most datasets while achieving comparable results for FGVC-Aicraft dataset. This indicates that our method is effective for fine-grained category discovery. On the domain adaptation classification front, our method shows the best results across all subsets '\textit{All}', '\textit{Old}', and '\textit{New}' on the DomainNet dataset, which indicates that our method is much more robust to distribution shift than standard ImageNet pre-trained models.

\subsection{Analysis}
\begin{table}
\caption{
Image only versus Image and Text clustering accuracy with different image encoders.
}\label{table:ablation}
\centering
\renewcommand{\arraystretch}{1.2}
\resizebox{1.0\linewidth}{!}{
\begin{tabular}{@{\extracolsep{0pt}}cccccc@{}}
\toprule
Dataset & Image Encoder & Knowledge & All & Old & New \\
\midrule
CIFAR-100 & GCD & N & 73.0 & 76.2 & 66.5\\
CIFAR-100 & GCD & Y & 75.9 & 79.7 & 67.3\\
CIFAR-100 & CLIP & N & 84.2 & 83.1 & 82.3\\
CIFAR-100 & CLIP & Y & 85.2 & 85.0 & 85.6\\
Stanford Cars & GCD & N & 39.0 & 57.6 & 29.9\\
Stanford Cars & GCD & Y & 41.1 & 60.0 & 33.5\\
Stanford Cars & CLIP & N & 62.8 & 85.2 & 52.0\\
Stanford Cars & CLIP & Y & 70.6 & 88.2 & 62.2\\
Sketch & GCD & N & 30.2 & 46.4 & 24.3\\
Sketch & GCD & Y & 30.9 & 48.3 & 25.9\\
Sketch & CLIP & N & 52.7 & 74.2 & 43.7\\
Sketch & CLIP & Y & 55.2 & 75.5 & 47.4\\
\midrule
Average & GCD & N & 47.4 & 60.1 & 40.2\\
Average & GCD & Y & 49.3 & 62.7 & 42.2\\
Average & CLIP & N & 66.6 & 80.8 & 59.3\\
Average & CLIP & Y & \textbf{70.3} & \textbf{82.9} & \textbf{65.1}\\
\bottomrule
\end{tabular}
}
\end{table}
We analyze the contribution of certain aspects of our methodology through a rigorous ablation study. Specifically, we highlight the significance of the following components of the approach: whether language supervision can result in vision models with transferable representation versus classic image-only models, the effect of the number of texts \textit{k} retrieved per image on the accuracy of the model, retrieved text quality, and CLIP image encoder ViT backbone with and without finetuning. 

\begin{table}
\caption{
Accuracy of the model using different knowledge databases as a source of text descriptions
}\label{table:knowledge_db}
\centering
\renewcommand{\arraystretch}{1.2}
\resizebox{0.8\linewidth}{!}{
\begin{tabular}{@{\extracolsep{0pt}}ccccc@{}}
\toprule
Dataset & Knowledge DB & All & Old & New \\
\midrule
CIFAR-100 & CC-12M & 85.9 & 85.0 & 88.1\\
CIFAR-100 & CC-3M & 82.8 & 82.6 & 83.2\\
CIFAR-100 & MSCOCO & 85.1 & 85.5 & 84.2\\
CIFAR-100 & LAION-400M & 82.0 & 82.6 & 80.8\\
CIFAR-100 & LAION-5B & 82.5 & 83.4 & 80.6\\
Stanford Cars & CC-12 & 70.9 & 89.3 & 62.0\\
Stanford Cars & CC-3M & 63.8 & 85.1 & 53.5\\
Stanford Cars & MSCOCO & 62.4 & 85.5 & 51.2\\
Stanford Cars & LAION-400M & 66.1 & 86.7 & 56.1\\
Stanford Cars & LAION-5B & 71.2 & 89.4 & 64.5\\
Sketch & CC-12 & 54.7 & 74.6 & 47.4\\
Sketch & CC-3M & 55.2 & 76.8 & 47.6\\
Sketch & MSCOCO & 55.2 & 78.2 & 47.2\\
Sketch & LAION-400M & 53.8 & 76.1 & 45.3\\
Sketch & LAION-5B & 54.6 & 77.3 & 46.9\\
\midrule
Average & CC-12M & \textbf{70.5} & \textbf{83.0} & \textbf{65.8}\\
Average & CC-3M & 67.3 & 81.5 & 61.4\\
Average & MSCOCO & 67.7 & 83.1 & 60.9\\
Average & LAION-400M & 67.3 & 81.8 & 60.7\\
Average & LAION-5B & 69.4 & 83.0 & 64.0\\
\bottomrule
\end{tabular}
}
\end{table}
\noindent \textbf{How important is language supervision in this setting?} Table \ref{table:ablation} shows the effect of language on the clustering task. The \textit{Image Encoder} column represents different types of vision transformer backbones. GCD is a finetuned ViT-B-16 backbone with DINO \cite{Caron2021EmergingPI} pre-trained weights from GCD  \cite{Vaze2022GeneralizedCD} and CLIP \cite{Radford2021LearningTV} is a finetuned pre-trained ViT-B-16 backbone. The \textit{Knowledge} columns indicate whether we are clustering vision-only features or vision and text features combined. We record the accuracy of the model across all categories, All, Old, and New for three datasets, then average them for each combination of dataset, image encoder, and knowledge. As shown, the results indicate that CLIP image and text outperform image-only by a large margin, confirming that language does help in this setting compared to image-only models. We note that while using CLIP as an encoder without retrieval is an extremely strong baseline, our retrieval mechanism further improves performance by significant margins e.g. almost 4\% on All and almost 6\% specifically on Old. 

\noindent \textbf{How important is the descriptiveness of retrieved captions?} Text descriptions in typical datasets can vary in terms of how they relate to the image. Ideally, we want to encode salient objects in the image that are meaningful in representation learning for object recognition tasks. The learned representations for contrastive models are governed by the text transformer (captions for CLIP), suggesting that text descriptions that describe the contents of a scene in an image will improve transferability in the CLIP model. We verify this hypothesis and quantify the \textit{descriptiveness} of a caption using multiple caption data sources. We perform top-4 cross-modal retrieval from Conceptual Captions (3M) \cite{Sharma2018ConceptualCA}, Conceptual Captions (12M) \cite{Changpinyo2021Conceptual1P}, and COCO \cite{Lin2014MicrosoftCC}, and LION \cite{Schuhmann2021LAION400MOD}, then record the accuracy of the model for each data corpus on \textit{All}, \textit{Old}, and \textit{New} subsets averaged for each  knowledge database.

\looseness=-1 Table \ref{table:knowledge_db} shows the results of the model on three datasets CIFAR100, Stanford Cars, and DomainNet(Sketch). Previous work in linguistics has shown that captions that are descriptive (meant to replace an image) are different from those that are complementary or give additional information and context. Contrary to LAION and Conceptual Captions (12M) which usually contain information complementary to the image,  Conceptual Captions (3M) and MS COCO are more descriptive due to the strict annotation process. We use a score given by CLIP of a caption matching its corresponding image in our cross-modal retrieval, and according to the  results, the hypothesis does not align with our subjective assessment, at least for the datasets tested. We posit that the descriptiveness of the captions retrieved from a corpus and the size of the knowledge database, as well as the diversity of captions, all play a role.

\textbf{How many captions do we need to retrieve for each image?} We probe how the variability of captions within a caption database affects our model transfer capabilities. There are a number of ways to annotate an image as shown in Figure \ref{fig:figure3}. In each corpus, captions vary in terms of how an object is described e.g. "train" or "railcar", and which part of the image the focus is on, e.g. "cloud" or "bird". The focus, lexical, and style variation in captioning could confuse the model and make it push image-text pairs apart instead of pulling them together. We examine the sensitivity of our model to the number of captions per image (top-k), averaging accuracy across three datasets, CIFAR100, Stanford Cars, and DomainNet (Sketch), and we chose to limit retrieval of captions to Conceptual Captions (12M)\cite{Changpinyo2021Conceptual1P}.

\looseness=-1 Figure \ref{fig:figure3} suggests that variability in dataset captions can hurt the accuracy of the model. They suggest that some of the captions might not contain useful information making the model accuracy plateau or even reduce after a certain number. 

\begin{table}
\caption{
Results for finetuned CLIP vs. not finetuned
}\label{table:finetuning}
\centering
\renewcommand{\arraystretch}{1.2}
\resizebox{1.0\linewidth}{!}{
\begin{tabular}{@{\extracolsep{0pt}}ccccc@{}}
\toprule
Dataset & Finetuned CLIP & All & Old & New \\
\midrule
CIFAR-100 & N & 68.7 & 71.1 & 63.7\\
CIFAR-100 & Y & 85.2 & 85.0 & 85.6\\
Stanford Cars & N & 65.8 & 78.9 & 59.5\\
Stanford Cars & Y & 70.6 & 88.2 & 62.2\\
Sketch & N & 51.6 & 60.0 & 48.5\\
Sketch & Y & 55.2 & 75.5 & 47.4\\
\midrule
Average & N & 62.0 & 70.0 & 57.2\\
Average & Y & \textbf{70.3} & \textbf{82.9} & \textbf{65.1}\\
\bottomrule
\end{tabular}
}
\end{table}

\textbf{Does CLIP need finetuning?}
One of the most impressive aspects of the CLIP model is its performance in zero-shot learning, classifying objects it has never seen before, based on their descriptions in natural language. In this experiment, we probe CLIP's performance in the GCD setting without performing any finetuning. Table \ref{table:finetuning} shows our results for a CLIP model finetuned versus a model without finetuning on three datasets, CIFAR100, Stanford Cars, and Sketch with a finetuned CLIP outperforming a non-finetuned CLIP model. Recent studies have shown that CLIP finetuning might distort its pretrained representation leading to unsatisfactory performance, but our results show that it can be finetuned with the right hyperparameter choices, challenging the notion that CLIP is not suitable for finetuning.

\begin{figure*}[t]
     \centering
     \begin{subfigure}[b]{0.33\textwidth}
         \centering
         \includegraphics[width=\textwidth]{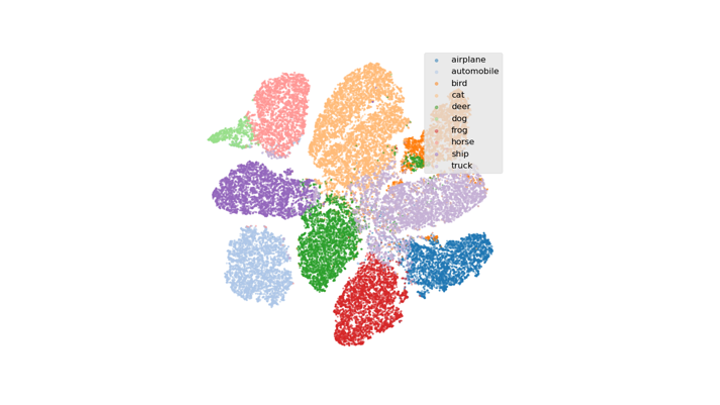}
         \caption{Image only feature visualization on CIFAR10 with t-SNE}
     \end{subfigure}
     \hfill
     \begin{subfigure}[b]{0.33\textwidth}
         \centering
         \includegraphics[width=\textwidth]{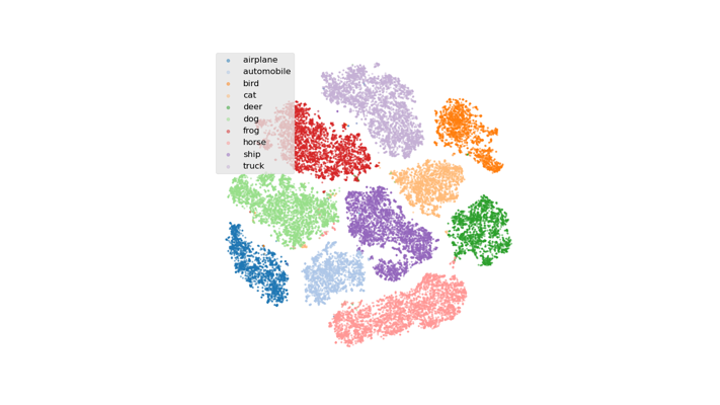}
         \caption{Image and Text feature visualization on CIFAR10 with t-SNE}
     \end{subfigure}
     \hfill
     \begin{subfigure}[b]{0.33\textwidth}
         \centering
         \includegraphics[width=\textwidth]{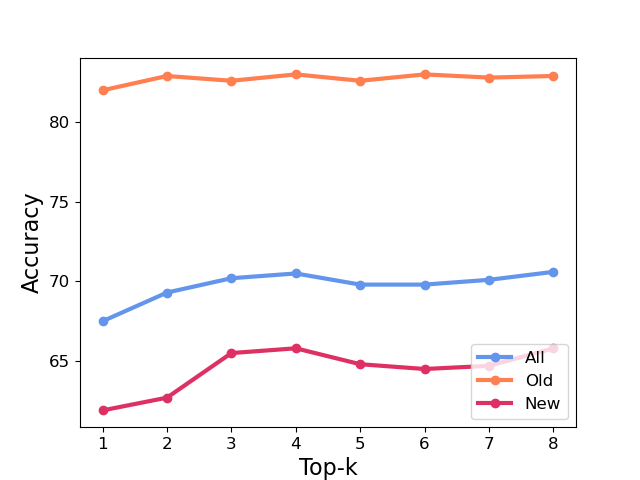}
         \caption{Sensitivity of the model to the \# of captions per image averaged across three datatsets.}
     \end{subfigure}
     \caption{}
    \label{fig:tsne_figure}
\end{figure*}

\subsection{Qualitative results} 
We  further show a t-SNE \cite{Maaten2014AcceleratingTU} projection of ViT CLIP image features and Image-Text features to visualize the feature spaces of CIFAR10 by transforming the features into two dimensions. In Figure \ref{fig:tsne_figure}, we show the clustered features of the unlabeled data and compared the results of our method for image-only features against image and text features. For image-only features, data points from the same class are generally projected close to each other, and they form clear clusters with some overlapping between classes. In contrast, the image-text features form clear clusters with some clear separation which are further distinguished when using text along with an image, further confirming the utility of language in this setting.
\label{sec:method}

\section{Conclusion}
In this paper, we propose to tackle the Generalized Category Discovery setting. With the recent advances in Vision-Language pertaining (VLP), we propose to use CLIP and take advantage of its multi-modality in two ways. First, we propose to leverage the CLIP image encoder, yielding an extremely strong baseline for GCD. Second, we propose a complementary novel retrieval-based augmentation, specifically retrieving textual context from a text corpus and jointly clustering the image and text embeddings.  We perform rigorous analysis demonstrating that our method  is well suited for this setting. 

We demonstrate significant quantitative improvements on four generic classifications, three fine-grained, and one domain adaptation datasets showing significant performance gains over previous methods. Importantly, we show that our two ways of leveraging CLIP are complementary and that both are necessary to achieve strong state-of-art results. There are a number of limitations and future work, including enhancing the retrieval process to improve the quality of the retrieved contextual knowledge. 
\label{sec:conclusion}

{\small
\bibliographystyle{ieee_fullname}

\begin{thebibliography}{10}\itemsep=-1pt

\bibitem{Arthur2007kmeansTA}
David Arthur and Sergei Vassilvitskii.
\newblock k-means++: the advantages of careful seeding.
\newblock In {\em SODA '07}, 2007.

\bibitem{Caron2021EmergingPI}
Mathilde Caron, Hugo Touvron, Ishan Misra, Herv'e J'egou, Julien Mairal, Piotr
  Bojanowski, and Armand Joulin.
\newblock Emerging properties in self-supervised vision transformers.
\newblock {\em 2021 IEEE/CVF International Conference on Computer Vision
  (ICCV)}, pages 9630--9640, 2021.

\bibitem{Chang2017DeepAI}
Jianlong Chang, Lingfeng Wang, Gaofeng Meng, Shiming Xiang, and Chunhong Pan.
\newblock Deep adaptive image clustering.
\newblock {\em 2017 IEEE International Conference on Computer Vision (ICCV)},
  pages 5880--5888, 2017.

\bibitem{Changpinyo2021Conceptual1P}
Soravit Changpinyo, Piyush~Kumar Sharma, Nan Ding, and Radu Soricut.
\newblock Conceptual 12m: Pushing web-scale image-text pre-training to
  recognize long-tail visual concepts.
\newblock {\em 2021 IEEE/CVF Conference on Computer Vision and Pattern
  Recognition (CVPR)}, pages 3557--3567, 2021.

\bibitem{Chapelle2006SemiSupervisedL}
Olivier Chapelle, Bernhard Schlkopf, and Alexander Zien.
\newblock Semi-supervised learning.
\newblock {\em IEEE Transactions on Neural Networks}, 20, 2006.

\bibitem{chen2020simple}
Ting Chen, Simon Kornblith, Mohammad Norouzi, and Geoffrey Hinton.
\newblock A simple framework for contrastive learning of visual
  representations.
\newblock In {\em International conference on machine learning}, pages
  1597--1607. PMLR, 2020.

\bibitem{Deng2009ImageNetAL}
Jia Deng, Wei Dong, Richard Socher, Li-Jia Li, K. Li, and Li Fei-Fei.
\newblock Imagenet: A large-scale hierarchical image database.
\newblock {\em 2009 IEEE Conference on Computer Vision and Pattern
  Recognition}, pages 248--255, 2009.

\bibitem{dosovitskiy2020image}
Alexey Dosovitskiy, Lucas Beyer, Alexander Kolesnikov, Dirk Weissenborn,
  Xiaohua Zhai, Thomas Unterthiner, Mostafa Dehghani, Matthias Minderer, Georg
  Heigold, Sylvain Gelly, et~al.
\newblock An image is worth 16x16 words: Transformers for image recognition at
  scale.
\newblock {\em arXiv preprint arXiv:2010.11929}, 2020.

\bibitem{Fei2022XConLW}
Yixin Fei, Zhongkai Zhao, Si~Xiao Yang, and Bingchen Zhao.
\newblock Xcon: Learning with experts for fine-grained category discovery.
\newblock {\em ArXiv}, abs/2208.01898, 2022.

\bibitem{Fini2021AUO}
Enrico Fini, E. Sangineto, St{\'e}phane Lathuili{\`e}re, Zhun Zhong, Moin Nabi,
  and Elisa Ricci.
\newblock A unified objective for novel class discovery.
\newblock {\em 2021 IEEE/CVF International Conference on Computer Vision
  (ICCV)}, pages 9264--9272, 2021.

\bibitem{Furst2021CLOOBMH}
Andreas Furst, Elisabeth Rumetshofer, Viet-Hung Tran, Hubert Ramsauer, Fei
  Tang, Johannes Lehner, David~P. Kreil, Michael Kopp, G{\"u}nter Klambauer,
  Angela Bitto-Nemling, and Sepp Hochreiter.
\newblock Cloob: Modern hopfield networks with infoloob outperform clip.
\newblock {\em ArXiv}, abs/2110.11316, 2021.

\bibitem{VanGansbeke2020LearningTC}
Wouter~Van Gansbeke, Simon Vandenhende, Stamatios Georgoulis, Marc Proesmans,
  and Luc~Van Gool.
\newblock Learning to classify images without labels.
\newblock {\em ArXiv}, abs/2005.12320, 2020.

\bibitem{Gong2014ImprovingIE}
Yunchao Gong, Liwei Wang, Micah Hodosh, J. Hockenmaier, and Svetlana Lazebnik.
\newblock Improving image-sentence embeddings using large weakly annotated
  photo collections.
\newblock In {\em ECCV}, 2014.

\bibitem{grill2020bootstrap}
Jean-Bastien Grill, Florian Strub, Florent Altch{\'e}, Corentin Tallec, Pierre
  Richemond, Elena Buchatskaya, Carl Doersch, Bernardo Avila~Pires, Zhaohan
  Guo, Mohammad Gheshlaghi~Azar, et~al.
\newblock Bootstrap your own latent-a new approach to self-supervised learning.
\newblock {\em Advances in neural information processing systems},
  33:21271--21284, 2020.

\bibitem{Han2020AutomaticallyDA}
K. Han, Sylvestre-Alvise Rebuffi, S{\'e}bastien Ehrhardt, Andrea Vedaldi, and
  Andrew Zisserman.
\newblock Automatically discovering and learning new visual categories with
  ranking statistics.
\newblock {\em ArXiv}, abs/2002.05714, 2020.

\bibitem{Han2022AutoNovelAD}
K. Han, Sylvestre-Alvise Rebuffi, S{\'e}bastien Ehrhardt, Andrea Vedaldi, and
  Andrew Zisserman.
\newblock Autonovel: Automatically discovering and learning novel visual
  categories.
\newblock {\em IEEE Transactions on Pattern Analysis and Machine Intelligence},
  44:6767--6781, 2022.

\bibitem{Han2019LearningTD}
K. Han, Andrea Vedaldi, and Andrew Zisserman.
\newblock Learning to discover novel visual categories via deep transfer
  clustering.
\newblock {\em 2019 IEEE/CVF International Conference on Computer Vision
  (ICCV)}, pages 8400--8408, 2019.

\bibitem{he2022masked}
Kaiming He, Xinlei Chen, Saining Xie, Yanghao Li, Piotr Doll{\'a}r, and Ross
  Girshick.
\newblock Masked autoencoders are scalable vision learners.
\newblock In {\em Proceedings of the IEEE/CVF Conference on Computer Vision and
  Pattern Recognition}, pages 16000--16009, 2022.

\bibitem{Hendrycks2019UsingSL}
Dan Hendrycks, Mantas Mazeika, Saurav Kadavath, and Dawn~Xiaodong Song.
\newblock Using self-supervised learning can improve model robustness and
  uncertainty.
\newblock In {\em NeurIPS}, 2019.

\bibitem{HodoshMicah2013FramingID}
HodoshMicah, YoungPeter, and HockenmaierJulia.
\newblock Framing image description as a ranking task.
\newblock {\em Journal of Artificial Intelligence Research}, 2013.

\bibitem{hsu2015neural}
Yen-Chang Hsu and Zsolt Kira.
\newblock Neural network-based clustering using pairwise constraints.
\newblock {\em arXiv preprint arXiv:1511.06321}, 2015.

\bibitem{Hsu2018LearningTC}
Yen-Chang Hsu, Zhaoyang Lv, and Zsolt Kira.
\newblock Learning to cluster in order to transfer across domains and tasks.
\newblock {\em ArXiv}, abs/1711.10125, 2018.

\bibitem{Hsu2019MulticlassCW}
Yen-Chang Hsu, Zhaoyang Lv, Joel Schlosser, Phillip Odom, and Zsolt Kira.
\newblock Multi-class classification without multi-class labels.
\newblock {\em ArXiv}, abs/1901.00544, 2019.

\bibitem{Jia2021ScalingUV}
Chao Jia, Yinfei Yang, Ye Xia, Yi-Ting Chen, Zarana Parekh, Hieu Pham, Quoc~V.
  Le, Yun-Neuralan Sung, Zhen Li, and Tom Duerig.
\newblock Scaling up visual and vision-language representation learning with
  noisy text supervision.
\newblock In {\em ICML}, 2021.

\bibitem{Khalid2022RODDAS}
Umar Khalid, Ashkan Esmaeili, Nazmul Karim, and Nazanin Rahnavard.
\newblock Rodd: A self-supervised approach for robust out-of-distribution
  detection.
\newblock {\em 2022 IEEE/CVF Conference on Computer Vision and Pattern
  Recognition Workshops (CVPRW)}, pages 163--170, 2022.

\bibitem{Krause20133DOR}
Jonathan Krause, Michael Stark, Jia Deng, and Li Fei-Fei.
\newblock 3d object representations for fine-grained categorization.
\newblock {\em 2013 IEEE International Conference on Computer Vision
  Workshops}, pages 554--561, 2013.

\bibitem{Krizhevsky2009LearningML}
Alex Krizhevsky.
\newblock Learning multiple layers of features from tiny images.
\newblock 2009.

\bibitem{Kuhn2010TheHM}
Harold~W. Kuhn.
\newblock The hungarian method for the assignment problem.
\newblock {\em Naval Research Logistics (NRL)}, 52, 2010.

\bibitem{kuo2022beyond}
Chia-Wen Kuo and Zsolt Kira.
\newblock Beyond a pre-trained object detector: Cross-modal textual and visual
  context for image captioning.
\newblock In {\em Proceedings of the IEEE/CVF Conference on Computer Vision and
  Pattern Recognition}, pages 17969--17979, 2022.

\bibitem{Li2022SupervisionEE}
Yangguang Li, Feng Liang, Lichen Zhao, Yufeng Cui, Wanli Ouyang, Jing Shao,
  Fengwei Yu, and Junjie Yan.
\newblock Supervision exists everywhere: A data efficient contrastive
  language-image pre-training paradigm.
\newblock {\em ArXiv}, abs/2110.05208, 2022.

\bibitem{Lin2014MicrosoftCC}
Tsung-Yi Lin, Michael Maire, Serge~J. Belongie, James Hays, Pietro Perona, Deva
  Ramanan, Piotr Doll{\'a}r, and C.~Lawrence Zitnick.
\newblock Microsoft coco: Common objects in context.
\newblock In {\em ECCV}, 2014.

\bibitem{MacQueen1967SomeMF}
J. MacQueen.
\newblock Some methods for classification and analysis of multivariate
  observations.
\newblock 1967.

\bibitem{Maji2013FineGrainedVC}
Subhransu Maji, Esa Rahtu, Juho Kannala, Matthew~B. Blaschko, and Andrea
  Vedaldi.
\newblock Fine-grained visual classification of aircraft.
\newblock {\em ArXiv}, abs/1306.5151, 2013.

\bibitem{Nilsback08}
Maria-Elena Nilsback and Andrew Zisserman.
\newblock Automated flower classification over a large number of classes.
\newblock In {\em Indian Conference on Computer Vision, Graphics and Image
  Processing}, Dec 2008.

\bibitem{Oza2019C2AECC}
Poojan Oza and Vishal~M. Patel.
\newblock C2ae: Class conditioned auto-encoder for open-set recognition.
\newblock {\em 2019 IEEE/CVF Conference on Computer Vision and Pattern
  Recognition (CVPR)}, pages 2302--2311, 2019.

\bibitem{peng2019moment}
Xingchao Peng, Qinxun Bai, Xide Xia, Zijun Huang, Kate Saenko, and Bo Wang.
\newblock Moment matching for multi-source domain adaptation.
\newblock In {\em Proceedings of the IEEE International Conference on Computer
  Vision}, pages 1406--1415, 2019.

\bibitem{Radford2021LearningTV}
Alec Radford, Jong~Wook Kim, Chris Hallacy, Aditya Ramesh, Gabriel Goh,
  Sandhini Agarwal, Girish Sastry, Amanda Askell, Pamela Mishkin, Jack Clark,
  Gretchen Krueger, and Ilya Sutskever.
\newblock Learning transferable visual models from natural language
  supervision.
\newblock In {\em ICML}, 2021.

\bibitem{Rebuffi2021LSDCLS}
Sylvestre-Alvise Rebuffi, S{\'e}bastien Ehrhardt, K. Han, Andrea Vedaldi, and
  Andrew Zisserman.
\newblock Lsd-c: Linearly separable deep clusters.
\newblock {\em 2021 IEEE/CVF International Conference on Computer Vision
  Workshops (ICCVW)}, pages 1038--1046, 2021.

\bibitem{Schuhmann2021LAION400MOD}
Christoph Schuhmann, Richard Vencu, Romain Beaumont, Robert Kaczmarczyk,
  Clayton Mullis, Aarush Katta, Theo Coombes, Jenia Jitsev, and Aran
  Komatsuzaki.
\newblock Laion-400m: Open dataset of clip-filtered 400 million image-text
  pairs.
\newblock {\em ArXiv}, abs/2111.02114, 2021.

\bibitem{Sharma2018ConceptualCA}
Piyush Sharma, Nan Ding, Sebastian Goodman, and Radu Soricut.
\newblock Conceptual captions: A cleaned, hypernymed, image alt-text dataset
  for automatic image captioning.
\newblock In {\em ACL}, 2018.

\bibitem{Sohn2020FixMatchSS}
Kihyuk Sohn, David Berthelot, Chun-Liang Li, Zizhao Zhang, Nicholas Carlini,
  Ekin~Dogus Cubuk, Alexey Kurakin, Han Zhang, and Colin Raffel.
\newblock Fixmatch: Simplifying semi-supervised learning with consistency and
  confidence.
\newblock {\em ArXiv}, abs/2001.07685, 2020.

\bibitem{Maaten2014AcceleratingTU}
Laurens van~der Maaten.
\newblock Accelerating t-sne using tree-based algorithms.
\newblock {\em J. Mach. Learn. Res.}, 15:3221--3245, 2014.

\bibitem{Vaze2022GeneralizedCD}
Sagar Vaze, K. Han, Andrea Vedaldi, and Andrew Zisserman.
\newblock Generalized category discovery.
\newblock {\em 2022 IEEE/CVF Conference on Computer Vision and Pattern
  Recognition (CVPR)}, pages 7482--7491, 2022.

\bibitem{Wah2011TheCB}
Catherine Wah, Steve Branson, Peter Welinder, Pietro Perona, and Serge~J.
  Belongie.
\newblock The caltech-ucsd birds-200-2011 dataset.
\newblock 2011.

\bibitem{Yang2017TowardsKS}
Bo Yang, Xiao Fu, N. Sidiropoulos, and Mingyi Hong.
\newblock Towards k-means-friendly spaces: Simultaneous deep learning and
  clustering.
\newblock In {\em ICML}, 2017.

\bibitem{ZelnikManor2004SelfTuningSC}
Lihi Zelnik-Manor and Pietro Perona.
\newblock Self-tuning spectral clustering.
\newblock In {\em NIPS}, 2004.

\bibitem{Zhong2021NeighborhoodCL}
Zhun Zhong, Enrico Fini, Subhankar Roy, Zhiming Luo, Elisa Ricci, and N. Sebe.
\newblock Neighborhood contrastive learning for novel class discovery.
\newblock {\em 2021 IEEE/CVF Conference on Computer Vision and Pattern
  Recognition (CVPR)}, pages 10862--10870, 2021.

\end{thebibliography}

}

\ifarxiv \clearpage \appendix
\addcontentsline{toc}{section}{Appendices}
\label{sec:appendix}
\twocolumn[%
   \begin{center}
     {\large\textbf{CLIP-GCD: Simple Language Guided Generalized Category Discovery}}\\
      \vspace{2ex}
      Appendices
   \end{center}
]

\section{CLIP ViT Backbone}
We investigate and compare CLIP ViT-B/16 versus ViT-L/14 (24 layers, a hidden size of 1024, and 307M parameters) to show the effect of a larger ViT model on the clustering task. 
\begin{table}[h]
\centering
\resizebox{\linewidth}{!}{%
\begin{tabular}{@{}lcclcclcc@{}}
\toprule
 & \multicolumn{2}{c}{All} &  & \multicolumn{2}{c}{Old} &  & \multicolumn{2}{c}{New} \\ \midrule
Dataset & ViT-B/16 & ViT-L/14 &  & ViT-B/16 & ViT-L/14 &  & ViT-B/16 & ViT-L/14 \\ \cmidrule(lr){2-3} \cmidrule(lr){5-6} \cmidrule(l){8-9} 
Stanford Cars & 70.6 & 75.2 &  & 88.2 & 91.3 &  & 62.2 & 67.4 \\
Sketch & 55.2 & 58.5 &  & 75.5 & 78.8 &  & 47.4 & 51.1 \\
CIFAR100 & 85.2 & 86.7 &  & 85.0 & 88.3 &  & 85.6 & 85.9 \\ \bottomrule
\end{tabular}%
}
\caption{Comparative results of our method accuracy of different ViT backbone sizes on Stanford Cars, DomainNet(Sketch), and CIFAR100 datasets}
\label{tab:vitbackbone}
\end{table} \\
We finetune the last block of ViT-L/14 transformer starting with a smaller learning rate of 4e-6 compared to ViT-B/16, decaying it over time using a cosine annealed schedule. We train the model for 100 epochs using batches of size 64.

Details are shown in table \ref{tab:vitbackbone}. ViT-L/14 performs better across different types of datasets, out-of-distribution, generic image recognition, and fine-grained. It outperforms ViT-B/16 by over 3\% aggregated over \textit{'All'}, \textit{'Old'}, and \textit{'New'} categories. It has been mentioned in \cite{Radford2021LearningTV} that zero-shot ImageNet validation set accuracy between ViT-L/14 and ViT-B/16 is over 7\% which validates our results.
 \fi

\end{document}